\pdfoutput=1

\documentclass[11pt]{article}

\usepackage[]{acl}

\usepackage{times}
\usepackage{latexsym}
\usepackage{tabularx}
\usepackage{multirow}
\usepackage{graphicx}

\usepackage{xcolor}         
\usepackage{colortbl}

\usepackage{amsmath}
\usepackage{amsfonts}
\usepackage{amssymb}
\DeclareMathOperator*{\argmax}{argmax}
\usepackage{enumitem}

\usepackage[T1]{fontenc}

\usepackage[utf8]{inputenc}

\usepackage{microtype}

\usepackage{inconsolata}

\usepackage{textcomp}
\usepackage{algorithm}
\usepackage{listings}{
\lstset{
  language=Python,
  upquote=true,
  showstringspaces=false,
  formfeed=newpage,
  tabsize=4,
  commentstyle=itshape,
  basicstyle=\scriptsize\ttfamily,
  morekeywords={models, lambda, forms}
}

\usepackage{todonotes}
\usepackage{xcolor}
\usepackage{microtype}
\usepackage{float}

\title{Evaluating Subword Tokenization:\\Alien Subword Composition and OOV Generalization Challenge}

\author{Khuyagbaatar Batsuren$^{1,2}$,~\;~Ekaterina Vylomova$^{1}$,~\;~Verna Dankers$^3$,\\~\;~\textbf{Tsetsuukhei Delgerbaatar$^{2}$,~\;~Omri Uzan$^{4}$,~\;~Yuval Pinter$^{4}$,~\and~Gábor Bella$^5$}\\
 $^1$University of Melbourne \quad $^2$National University of Mongolia \quad $^3$University of Edinburgh \\
$^4$Ben-Gurion University of the Negev \quad $^5$IMT Atlantique \\
 \texttt{khuyagbaatar.b@gmail.com}}

\begin{document}
\maketitle
\begin{abstract}
The popular subword tokenizers of current language models, such as \emph{Byte-Pair Encoding} (BPE), are known not to respect morpheme boundaries, which affects the downstream performance of the models. While many improved tokenization algorithms have been proposed, their evaluation and cross-comparison are still an open problem. As a solution, we propose a combined intrinsic--extrinsic evaluation framework for subword tokenization. Intrinsic evaluation is based on our new \emph{UniMorph Labeller} (umLabeller) tool that classifies a subword tokenization as either  \emph{morphological} or \emph{alien}. Extrinsic evaluation, in turn, is performed via the \emph{Out-of-Vocabulary Generalization Challenge~1.0} benchmark, which consists of three newly specified downstream text classification tasks.
Our empirical findings show that the accuracy of umLabeller is 98\%, and that, in all language models studied (including ALBERT, BERT, RoBERTa, and DeBERTa), alien tokenization leads to poorer generalizations compared to morphological tokenization for semantic compositionality of word meanings. 
\end{abstract}

\section{Introduction}
Subword tokenization is a fundamental preprocessing method in Natural Language Processing that segments 
words into subword units.
Popular subword tokenization methods, such as Byte Pair Encoding \cite[BPE;][]{sennrich2016neural} or Unigram Language Model  \cite[ULM;][]{kudo2018subword}, are adaptations of data compression algorithms that mainly rely on word character co-occurrence statistics 
in a given text corpus, rather than on human knowledge and understanding about word formations and morphology.
As a result, certain subword compositions produced by these tokenizers are not aligned with any semantic compositions as understood by humans (e.g.,~\textit{h \_iked} in the GPT-4 Tokenizer).
In the rest of the paper, we refer to such linguistically implausible subword compositions as \textit{alien} compositions.
In alien subword compositions, subwords are not recognized by us humans as meaningful units from which the overall word meaning is composed. For example, in case of \textit{j \_ogging} (segmentation produced by GPT-3 and RoBERTa tokenizers), no subword in this composition represents a meaning of \textit{jog}. This issue is well illustrated by adversarial attacks, as shown in Table~\ref{tab:adv}, to which current language models are vulnerable.

\begin{figure}[t]
\centering
\includegraphics[width=1\linewidth]{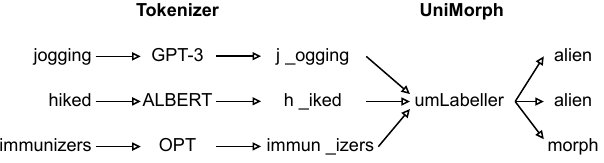}
\caption{umLabeller, the inspection tool for characterizing semantic compositionality of the subword-level tokenization into morph and alien.}
\label{fig:umlabeller_examples}
\end{figure}

Despite the success of subword tokenization in popular NLP applications (including machine translation, text generation, and text classification) and a wide range of practical \cite{mielke2021between} and cognitive studies \cite{beinborn2023analyzing}, evaluating subword tokenization algorithms is still an open problem for at least two reasons. Firstly, state-of-the-art evaluations in NLP models lack a unified set of criteria, as well as the underlying decision process, to verify the intrinsic correctness of a given subword tokenization. The second motivation is that little or no effort has gone into developing a standard extrinsic NLP benchmark to evaluate how tokenizers with different behaviors 
impact predictions of downstream tasks in NLP \cite{truong2024revisiting}.


\begin{table*}[ht]
\caption[]{\label{tab:adv} Adversarial attack example on Natural Language Inference with umLabeller. Here, we evaluated the RoBERTa-Large-MNLI\footnotemark model on the adversarial instances generated using the SNLI dataset \cite{bowman-etal-2015-large} by first detecting alien tokenizations with umLabeller and then swapping them with other alien tokenizations.}
\newcolumntype{R}{>{\raggedleft\arraybackslash}X}
\newcolumntype{C}{>{\centering\arraybackslash}X}
\newcolumntype{L}{>{\raggedright\arraybackslash}X}

\begin{tabularx}{\textwidth}{l|l|L|l}
                   & type      & input                                                                                               & prediction             \\
                   \hline
                   \hline
\multirow{2}{*}{1} & original    & A woman in orange sunglasses \textbf{\textit{jogs}}. A person is \textbf{\textit{jogging}}.                                             & entailment (0.988) \\
\cline{2-4}
                   & adversarial & A woman in orange sunglasses \textbf{\textit{embogs}}. A person is \textbf{\textit{shogging}}.                                          & entailment (0.836) \\
\hline
                   \hline
\multirow{2}{*}{\shortstack[l]{~\\~\\~\\~\\~\\~\\~\\~\\2}} & \multirow{1}{*}{\shortstack[l]{~\\~\\~\\original}}    & A male is \textbf{\textit{mowing}} the grass with a lawn \textbf{\textit{mower}}. A person is \textbf{\textit{mowing}} the grass with a lawn \textbf{\textit{mower}}.       & \multirow{1}{*}{\shortstack[l]{~\\~\\~\\entailment (0.992)}} \\
\cline{2-4}
                   & \multirow{1}{*}{\shortstack[l]{~\\~\\~\\adversarial}}  & A male is \textbf{\textit{undertowing}} the grass with a lawn \textbf{\textit{avower}}. A person is \textbf{\textit{cowing}} the grass with a lawn \textbf{\textit{bower}}. & \multirow{1}{*}{\shortstack[l]{~\\~\\~\\entailment (0.854)}}
                   \\ 
                   \hline
\end{tabularx}
\end{table*}

\begin{table*}[ht]
\centering
\caption{\label{tab:umlabeller} umLabeller with example inputs from the GPT and ALBERT tokenizers}
\newcolumntype{R}{>{\raggedleft\arraybackslash}X}
\newcolumntype{C}{>{\centering\arraybackslash}X}
\newcolumntype{L}{>{\raggedright\arraybackslash}X}

\begin{tabularx}{\textwidth}{l|l|Ll|ll}
Input            & UniMorph                      & GPT Tokenizer        & umLabeller & ALBERT Tokenizer        & umLabeller \\
\hline
\hline
jogging            & \_jog ing                     & \_j ogging               & alien     & \_jogging                 & vocab     \\
neutralised        & \_neuter al ise ed        & \_neutral ised           & morph     & \_neutral ised          & morph     \\
stepstones         & \_step stone s              & \_step stones            & morph     & \_steps tones           & alien     \\
clerking           & \_clerk ing                   & \_cler king              & alien     & \_clerk ing             & morph     \\
swappiness & \_swap y ness & \_sw appiness & alien     & \_swap pi ness & morph    
\end{tabularx}
\end{table*}

In this work, we first describe \textit{umLabeller}, a large-scale, high-quality characterization algorithm and tool for subword compositions.
\textit{umLabeller} is well adapted to the needs of both subword tokenization and morphological segmentations. The example usages of umLabeller are shown in \autoref{fig:umlabeller_examples} and \autoref{tab:umlabeller}.
Secondly, we present the OOV Generalization Challenge 1.0 Benchmark for evaluating subword tokenizations based on output from umLabeller.
This challenge contains three downstream text classification subtasks, and evaluates the compositional and morphological generalization abilities of language models under fully-generated covariate shifts between the fine-tuning and test stage. 


\footnotetext{\url{https://huggingface.co/FacebookAI/roberta-large-mnli}}





Our main contributions are as follows:
\begin{itemize}[leftmargin=*]
    \item umLabeller: an algorithm for classifying subword composition as either morphological or alien, implemented as an open-source tool, and covering over half a million English words;
    \item evaluation of the accuracy of umLabeller: we deem its accuracy of 98.0\% to be sufficient for the vast majority of NLP tasks;
    \item the Out-of-Vocabulary Generalization Challenge: a first-of-its-kind benchmark for evaluating subword tokenization in downstream NLP tasks; and
    \item empirical findings in our OOV Generalization Challenge showing that the alien compositions lead to poor generalization compared to morphological compositions and vocabulary words.
\end{itemize}

The paper is organized as follows. Section~\ref{sec:sota} presents the state-of-the-art subword tokenizations.
Section~\ref{sec:umlabeller} describes the umLabeller characterization algorithm for subword compositions, and Section~\ref{sec:challenge} presents methodologies on how each subtask dataset in OOV Generalization Challenge is generated from existing high-quality resources.
The umLabeller outputs are manually evaluated in Section~\ref{sec:validation}, and Section~\ref{sec:results} provides the experimental results of OOV Generalization Challenge.
Finally, we conclude the paper in section~\ref{sec:conclusion}.

\section{State-of-the-Art Subword Tokenization}
\label{sec:sota}

Tokenization has always been one of the traditional upstream tasks of NLP.
While most (Western) related work tends to associate tokens with \emph{words}, the latter is not a linguistically precise notion, and many languages (e.g.~written Chinese or Thai) do not split sentences into words.
Linguists consider the \emph{morpheme} as the atomic unit of meaning, and many languages distinguish between \emph{free morphemes} that can stand alone and \emph{bound morphemes} that are always used in combination with other morphemes. Nowadays, they also call it morph \cite{haspelmath2020morph} as the minimal linguistic form.
In traditional, symbolic approaches to computational linguistics, the task of \emph{morphological segmentation} means finding boundaries that delimit both free and bound morphemes.
This operation simultaneously restitutes the canonical forms of morphemes, as these may be altered through the composition of morphemes, e.g.~via \emph{allomorphs}.
Thus, the word \emph{copywriters} is morphologically segmented into the root words \emph{copy} and \emph{write} (of which \emph{copywrite} is a compound), a derivational affix \emph{-er}, and an inflectional affix \emph{-s}. 

In more recent neural approaches to language processing, which includes the well-known large language models (e.g.,~Llama~\cite{touvron2023llama}, Alpaca~\cite{taori2023alpaca}, OPT \cite{zhang2022opt}, GPT-4~\cite{achiam2023gpt}, or Bloom~\cite{le2022bloom}), \emph{subword tokenization} is invariably applied as a preprocessing step.
The goal of subword tokenization is to reduce the vocabulary size of the language model while also avoiding out-of-vocabulary words.
Following the corpus-based, fully inductive ethos of statistical and neural NLP, the most popular subword tokenization algorithms~\cite{sennrich2016neural,kudo2018subword} rely solely on character co-occurrence frequencies in order to predict token boundaries as a form of optimal text compression.

Linguistic compositionality itself being a way of compressing the infinity of linguistic expressivity into a limited lexicon of components; the hypothesis behind the application of mathematical compression methods to tokenization is that over large corpora, subword token boundaries coincide well with morpheme boundaries.
Practice shows, however, that this is only partially the case: firstly, purely frequency-based methods will not segment composed words that are themselves very frequent (e.g.~\emph{going}), nor will they analyze exceptional cases (e.g.~\emph{went}).
Secondly, these methods tend to conflate identical character strings into identical subword tokens:
they identify \emph{-iked} as a frequent subword due to its appearance in \textit{hiked}, \textit{biked}, and \textit{liked}, or \emph{-ogging} in \textit{jogging}, \textit{sogging}, and \textit{flogging}.
A third major difference between morphological analysis and subword tokenization is that the latter segments on the surface level and does not intend to restitute canonical forms of morphemes: \emph{angrily} will be split as \emph{angri|ly}, and the stem \emph{angry} is not identified.
Thus, morphology-unaware subword tokenization is prone to providing suboptimal results for downstream language understanding: on the one hand, unrelated character sequences are identified as mutual subwords, and on the other hand, the same morphemes having different surface forms are analyzed as unrelated.
For morphologically-rich languages, such phenomena are even more problematic~\cite{zhu2019importance}. 

\paragraph{Morphology-aware tokenization.}
An increasing number of efforts have attempted to improve on subword tokenization by applying morphology-aware tokenization methods to neural language models.
\emph{Canonical segmentation}~\cite{cotterell2016joint,kann-etal-2016-neural} and \emph{morphological segmentation}~\cite{girrbach-2022-sigmorphon,peters-martins-2022-beyond,bhandari2024using} focus on restoring the standard forms of morphemes in addition to computing surface segmentation.
Other efforts focus on preserving specific types of morphemes: derivation~\cite{hofmann2021superbizarre}, inflection~\cite{tan2020mind}, compounding~\cite{minixhofer2023compoundpiece}, or all of the above~\cite{hofmann2022embarrassingly,jabbar2023morphpiece}.
Some of these methods are limited to finding two-token subword compositions~\cite{hofmann2021superbizarre,yehezkel2023incorporating}. 

\paragraph{Evaluation of tokenization.}
Due to their varying goals and coverage, cross-comparison of these tokenization methods requires a coherent and unified evaluation methodology. In the literature, tokenization has been evaluated in two fundamental ways: intrinsically and extrinsically.

Intrinsic evaluation quantifies the output quality of subword tokenization by computing morphological alignment~\cite{uzan2024greed}, cognitive plausibility~\cite{beinborn2023analyzing,nair2023words,srivastava2024were}, or compression efficiency~\cite{zouhar2023tokenization,goldman2024unpacking}.
State-of-the-art morphological evaluation methods are limited in their coverage: (a)~many works suppose a fixed number of components (e.g.~two)~\cite{gow2022improving,hofmann2022embarrassingly}; (b)~they only recognise prefixes or suffixes~\cite{hofmann2021superbizarre,jacobs2022lost}; and (c)~they are limited to specific types of morphemes: derivation~\cite{beinborn2023analyzing,hofmann2022embarrassingly} or compounds~\cite{minixhofer2023compoundpiece}. 
Our four-class sequence labeling of tokens into \emph{vocabulary}, \emph{morphological}, \emph{alien}, and \emph{unknown}, defined in Section~\ref{sec:umlabeller}, provides an easy-to-understand and easy-to-use framework for evaluating a diversity of tokenization methods.
Our implementation via the umLabeller algorithm and tool is not affected by any of the limitations above: it can label words and subwords belonging to any morpheme type and at any level of granularity.
Therefore, umLabeller can be used as an intrinsic evaluation tool for any tokenizer.

Extrinsic evaluation focuses on how differences in tokenization impact neural model performance in downstream NLP tasks, e.g.,~machine translation~\cite{gowda2020finding,he2020dynamic,park2020empirical,zouhar2023tokenization},~question answering~\cite{sun2023tokenization,singh2024tokenization}, or text classification~\cite{bostrom-durrett-2020-byte,jabbar2023morphpiece,schmidt2024tokenization,goldman2024unpacking}.
However, the impact of tokenization on NLP tasks is often indirect and thus not easy to measure robustly: even if a word is incorrectly split, a successful prediction may still be made by the language model based on other tokens in context. 
Out-of-context tasks that depend on individual input words provide a more direct evaluation, so tasks with both context and individual input word was desirable for the evaluation.
The only such measure we found is the Word-in-Context (WiC) task \cite{pilehvar2019wic}. This task takes three elements in input: one word and two contextual sentences, and the expected prediction label is whether the word has the same meaning in both contexts. However, 80\% of all words of interest in the WiC test splits are vocabulary words by the commonly used language models and the remaining OOV words are limited to one to two hundred test instances.   

Our contribution with respect to extrinsic evaluation is a new evaluation method that we call the \emph{Out-of-Vocabulary Generalization Challenge}.
It consists of three downstream evaluation tasks, as presented in Section~\ref{sec:challenge}.

\section{umLabeller: UniMorph Labeller}
\label{sec:umlabeller}

\textsc{umLabeller}\footnote{https://github.com/unimorph/umLabeller} is the inspection tool for characterizing the subword-level composition of words, based on morphological information retrieved from UniMorph \cite{mccarthy-etal-2020-unimorph,batsuren-etal-2022-unimorph}. 
Let tokenizer $\mathbb{T}$ segment a given word $w$ into a sequence of subwords $s = (s_1,...,s_n)\,\,|\,\, \forall i\,\, s_i \in \mathcal{V}$ where $\mathcal{V}$ is a vocabulary list of subwords and $n$ is the length of the subword sequence. Given~$w$ and~$s$, $\textrm{umLabeller}$ assigns a label based on how $s$ is morphologically composed and aligned with UniMorph morpheme segmentations. The output label has four categories: \emph{vocab}, \emph{alien}, \emph{morph}, or \emph{n/a}:
\begin{itemize}[leftmargin=*]
    \item \textbf{\textit{vocabulary subword}}: the given word $w$ is a subword in the vocabulary as $w \in \mathcal{V}$;
    \item \textbf{\textit{alien composition}}: the given subword sequence $s$ is an \textit{alien subword composition} if we find at least two subwords $s_i$ and $s_j$ in $s$ that are not meaningful with respect to the meaning of $w$;
    \item \textbf{\textit{morphological composition}}: the subword sequence $s$ is morphological if it is neither a vocabulary nor an alien subword composition;
    \item \textbf{\textit{n/a}}: UniMorph has no information on the word. 
\end{itemize}



\subsection{Morphological Resources}
Below we describe how the umLabeller algorithm exploits morphological resources for its predictions. The main intuition behind using morphological resources is that morphological segmentation provides information about the compositional meaning of words \cite{vylomova2018compositional}.

\paragraph{UniMorph.} One of the state-of-the-art morphological databases in both NLP and CL is UniMorph \cite{sylak2015language,kirov2016very,kirov2018unimorph,mccarthy2020unimorph}, a morphosyntactic feature annotation of inflectional morphology in hundreds of languages, and the most recent version \cite{batsuren-etal-2022-unimorph} extended to morphological segmentation of inflectional and derivational morphology. For instance, the English word \texttt{motivated} is analyzed by UniMorph as follows:
\begin{center}
    \{\texttt{motivated} | infl | \texttt{motivate} | \texttt{-ed} | \texttt{\#V;PST}\}
    \{\texttt{motivate} | deri | \texttt{motive} | \texttt{-ate}\ | \texttt{\#N$\rightarrow$V}\} \\
\end{center}
from which the final segmentation \texttt{\_motive ate ed} is obtained. 

Subword tokenization methods, however, do not aim at full morphological segmentation but rather at obtaining a minimal number of subwords. For this reason, from the morphemes provided by UniMorph we produce all possible subword combinations. We first compute all possible morpheme n-grams for a given word. For the above example:
\begin{center}
    (\texttt{\_motive, ate, ed}), (\texttt{\_motive, ate}), \\(\texttt{ate, ed}), (\texttt{\_motive}), (\texttt{ate}), (\texttt{ed})
\end{center}

\noindent Then, we merge every bigram into a unigram by retrieving and inferring the target morpheme from UniMorph (taking morphological composition rules into account). During this process, if we have a new unigram 
(e.g., \texttt{ate + ed} $\rightarrow$ \texttt{ated}), we add it to the unigram list and generate all possible new n-grams (e.g.,~(\texttt{\_motive}, \texttt{ated})) with this unigram. This process continues until there is no new unigram. Finally, for all these n-grams (except for unigrams), we create a \textit{morphological merge list} by retrieving the target word or morpheme from UniMorph. For the current example, we create the following merge list:
\begin{center}
    (\texttt{\_motive, ate, ed}) $\rightarrow$ \texttt{\_motivated},\\
    (\texttt{\_motivate, ed}) $\rightarrow$ \texttt{\_motivated},\\
    (\texttt{\_motive, ated}) $\rightarrow$ \texttt{\_motivated},\\
    (\texttt{\_motive, ate}) $\rightarrow$ \texttt{\_motivate}, \\ (\texttt{ate, ed}) $\rightarrow$ \texttt{ated}\\
\end{center}

\paragraph{Compound morphology.} Our second resource is the SIGMORPHON-2022 Shared Task on morphological segmentation that provides compound segmentation data with inflectional and derivational affixes, derived from UniMorph 4.0 and MorphyNet \cite{batsuren2021morphynet}. For instance, the English word \texttt{copywriters} is segmented as follows:
\begin{center}
    \{\texttt{\_copywriters} | \texttt{\_copy write er s}\}
\end{center}
This piece of compound data complements the UniMorph segmentation and is fully compatible with our approach of generating the morphological merge list, as this resource linked with UniMorph~4.0.


\subsection{Algorithm}
The umLabeller algorithm described below relies on the morphological merge list computed in the previous section. 
Given a word $w$ and its subword sequence $s$, it implements the following classification rules:
\begin{equation*}
\textrm{umLabeller}(w,s) = \left\{\begin{matrix}
\textrm{vocab} & w \in \mathcal{V}\\ 
\textrm{n/a} & w \notin \textsc{UniMorph} \\ 
\textrm{alien} & \mu(w,s) < n-1 \\ 
\textrm{morph} & \mu(w,s) \geqslant n-1
\end{matrix}\right.
\end{equation*}
where $\mu$ is the number of subword tokens in~$s$ that can be found in the morphological merge list (and thus are deemed to be morphologically correct). Allowing for a tolerance of one missing element (hence $\mu < n-1$ and not $\mu <n$) is motivated by the possibility of root words not included in UniMorph, and by possible phonetic or orthographic alterations due to composition. For example, for the word \emph{theorizing}, the sequence \verb+_theor iz ing+ is classified by umLabeller as morphological because \texttt{izing} is included in the merge list computed in the previous section, and \texttt{theor} is the only token missing from it.
While we cannot formally prove the $\mu$~rule with respect to English morphology, we have strong empirical evidence for it: through all of our experiments, including our evaluations, we have not found a single counterexample, i.e.,~a morphologically correct composition that would contain two or more missing merge list entries.

The full algorithm is explained as follows:
\begin{itemize}[leftmargin=*]
    \item The algorithm first checks whether the vocabulary $\mathcal{V}$ contains the word $w$: if yes, it returns the label \emph{vocab}.
    \item It then verifies whether UniMorph contains any information about the word $w$. If no, it returns the label \emph{n/a}.
    \item The algorithm estimates $\mu(w,s)$, the number of morphological subwords in the subword sequence~$s$ by aligning all possible morphological merges of length~$n$, retrieved from UniMorph, as follows:
\begin{equation}
\small
\mu(w,s) = \argmax_{m \in \textsc{um}(w,n)}\sum_{i=1}^{n}|s_i=m_i|
\end{equation} 
where \textsc{um}($w$,$n$) returns a list of morphological merges for the word $w$ and its length equal to $n$, and $m=(m_1,...,m_n)$ is a sequence of morphological morphemes retrieved from UniMorph, morphological merges of $m$ is the word $w$. For instance, given the word \texttt{motivated} and the subword sequence length $n=2$, $\textsc{um}(w,n)$ is a list with two morphological merges as [(\texttt{\_motivate, ed}), (\texttt{\_motive, ated})]. 

\item if the number of morphological subwords, $\mu(w,s)$, is greater than or equal to $n$-1, $w$ is considered a morphological composition. Otherwise, it is labelled as \emph{alien}.
\end{itemize}




We provide a minimal Python implementation in Appendix \ref{app:algorithm}.

\begin{figure}[t]
\centering
\includegraphics[width=1\linewidth]{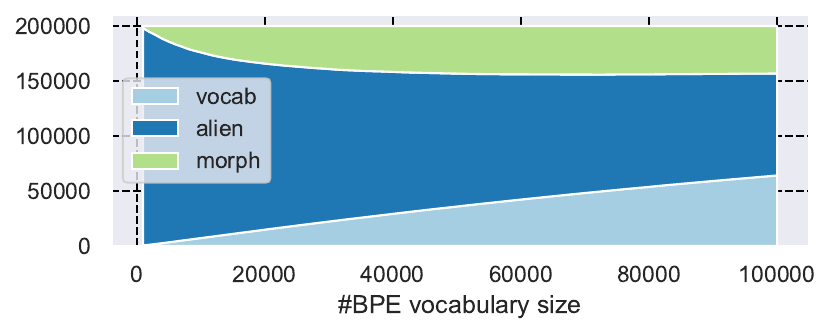}
\caption{umLabeller category distribution across BPE vocabulary sizes}
\label{fig:bpe-dist}
\end{figure}

\begin{table*}[ht]
\small
\centering
\caption{\label{tab:WaD} WaD dataset examples. The original word represents the corresponding definition in English WordNet.}
\newcolumntype{R}{>{\raggedleft\arraybackslash}X}
\newcolumntype{C}{>{\centering\arraybackslash}X}
\newcolumntype{L}{>{\raggedright\arraybackslash}X}
\begin{tabularx}{0.9\textwidth}{l|Ll|C|c}
 & \multicolumn{2}{c|}{input}     & output                                      &  \\

\hline
\hline
split & word     & definition                                      & label & original word  \\
\hline
\hline
\multirow{2}{*}{train}  & clerking & the activity of recording business transactions & true                & -                 \\ & ammo     & alternatively placed in genus Martynia          & false                      & martynia arenaria \\
\hline
\hline
\multirow{2}{*}{dev}  & enforced & forced or compelled or put in force & true                & -                 \\
& snowline & a fishing line managed principally by hand      & false               & handline          \\
\hline
\hline
\multirow{2}{*}{test}   & ovum & the female reproductive cell; the female gamete                          & true         & -          \\
 & mouther  & a wind from the south                           & false           & souther          \\
\end{tabularx}
\end{table*}

\begin{table}[t]
\small
\newcolumntype{R}{>{\raggedleft\arraybackslash}X}
\newcolumntype{C}{>{\centering\arraybackslash}X}
\newcolumntype{L}{>{\raggedright\arraybackslash}X}
\centering
\caption{\label{tab:WaM_example} WaM dataset examples.}
\begin{tabularx}{\linewidth}{l|CC|c}

& \multicolumn{2}{c|}{input}  & output             \\
\hline
\hline
\# & word         & morphology & label \\
\hline
\hline
1 & leaderboard  & derivation & true     \\
2 & overpressing & compound   & false     \\
3 & coteaches    & inflection & true     \\
4 & sharemarkets & derivation & false     \\
5 & untold       & inflection & true    \\

\hline
\end{tabularx}
\end{table}

\begin{table}[t]
\small
\newcolumntype{R}{>{\raggedleft\arraybackslash}X}
\newcolumntype{C}{>{\centering\arraybackslash}X}
\newcolumntype{L}{>{\raggedright\arraybackslash}X}
\centering
\caption{\label{tab:WaW_example} WaW dataset examples.}
\begin{tabularx}{\linewidth}{l|CC|c}

& \multicolumn{2}{c|}{input}  & output             \\
\hline
\hline
\# & $\textrm{word}_{a}$         & $\textrm{word}_{b}$  & label \\
\hline
\hline
1 & visitor  & traveler & true     \\
2 & shopper & earless   & false     \\
3 & photocopy    & mosaic & true     \\
4 & bleed & medicine & true     \\
5 & poorer       & proxy & false    \\

\hline
\end{tabularx}
\end{table}
 
\subsection{BPE Statistics on umLabeller}
In order to better understand state-of-the-art tokenizers, we trained 100 BPE tokenizers\footnote{We used the HuggingFace implementations from\\\url{https://github.com/huggingface/tokenizers}.} with increasing vocabulary sizes (from 1k to 100k with 1k increment). As the training corpus, we chose the Book Corpus \cite{zhu2015aligning} (4.5 GB of text data), which is used to pre-train BERT and RoBERTa. Then, we selected the 200,000 most frequent words based on the Book Corpus, applied the BPE tokenizers and labelled the resulting subword compositions using umLabeller. When selecting a vocabulary size, one wants to minimize the size while maximizing the morphological plausibility of word segmentations. These label distributions suggest that the optimal combination might be to use 40,000 to 50,000 tokens. When further increasing the vocabulary size, the number of words labeled as \emph{morph} no longer increases.

\begin{table*}[ht]
\small
\centering
\caption{\label{tab:wad_results}WaD test results by umLabeller categories. All language models are their corresponding large models. Tokenizer coverage represents the test split's coverage on the corresponding umLabeller category. All experiments trained on the best hyperparam sweep according to the validation results, averaged over five seeds.}
\newcolumntype{R}{>{\raggedleft\arraybackslash}X}
\newcolumntype{C}{>{\centering\arraybackslash}X}
\newcolumntype{L}{>{\raggedright\arraybackslash}X}
\begin{tabularx}{\textwidth}{l|cc|ccc|CCC|C}      & \multicolumn{2}{c|}{Tokenizer} & \multicolumn{3}{c|}{Test distribution} & \multicolumn{4}{c}{Accuracy}             \\
\hline
\hline
Model   & type & size & vocab & morph & alien & \multicolumn{1}{c}{vocab}       & \multicolumn{1}{c}{morph}        & \multicolumn{1}{c|}{alien}       & total \\
\hline
\hline
ALBERT  & ULM & 30k  & 27.0 & 44.6 & 28.5 & 93.3$\pm$0.3 & 71.4$\pm$2.7 & 67.2$\pm$2.9 & 76.2$\pm$1.9 \\
BERT    & BPE & 30k  & 26.9 & 37.9 & 35.2 & 95.3$\pm$0.4 & 77.0$\pm$1.4 & 72.4$\pm$1.8 & 80.4$\pm$1.0\\
RoBERTa & BPE & 50k  & 33.3 & 34.7 & 31.9 & 95.6$\pm$0.8 & 72.4$\pm$1.0 & 66.8$\pm$3.1 & 77.9$\pm$0.9\\
DeBERTa & ULM & 128k & 45.3 & 40.4 & 14.3 & 95.9$\pm$0.4 & 75.5$\pm$2.8 & 68.4$\pm$2.5 & 83.8$\pm$1.5\\
\hline
\end{tabularx}
\end{table*}

\begin{table*}[ht]
\small
\centering
\caption{\label{tab:WaM}WaM test results by umLabeller categories. All language models are their corresponding large models. Tokenizer coverage represents the test split's coverage on the corresponding umLabeller category. All experiments trained on the best hyperparam sweep according to the validation results, averaged over five seeds.}
\newcolumntype{R}{>{\raggedleft\arraybackslash}X}
\newcolumntype{C}{>{\centering\arraybackslash}X}
\newcolumntype{L}{>{\raggedright\arraybackslash}X}
\begin{tabularx}{\textwidth}{L|ccc|CCC|CCC|C}
        &  \multicolumn{3}{c|}{Test distribution} & \multicolumn{7}{c}{Accuracy}             \\
\hline
\hline
Model        & vocab & morph & alien & vocab & morph & alien & infl. & deri. & comp. & total \\
\hline
\hline
ALBERT  & 25.1  & 40.7  & 34.1  & 86.1$\pm$1.2 & 84.1$\pm$2.1 & 81.2$\pm$1.2 & 91.8$\pm$1.0 & 77.5$\pm$2.0     & 81.7$\pm$2.0    & 83.6$\pm$1.2 \\
BERT    & 26.9  & 27.7  & 45.2  & 88.5$\pm$1.0  & 86.2$\pm$0.8  & 82.3$\pm$0.5  & 89.7$\pm$0.9       & 80.8$\pm$0.5       & 84.8$\pm$0.5   & 85.0$\pm$0.4 \\
RoBERTa & 33.3  & 33.7  & 32.9  & 91.5$\pm$0.8  & 85.6$\pm$0.3 & 78.4$\pm$0.7 & 92.2$\pm$0.6     & 79.6$\pm$0.6      & 83.8$\pm$0.8   & 85.2$\pm$0.5 \\
DeBERTa & 36.3  & 42.1  & 21.6  & 91.8$\pm$0.6  & 87.5$\pm$1.1 & 84.8$\pm$1.0 & 95.4$\pm$0.4     & 82.6$\pm$1.7      & 87.3$\pm$0.9    & 88.4$\pm$0.8\\
\hline
\end{tabularx}
\end{table*}

\section{OOV Generalization Challenge}
\label{sec:challenge}
We present the OOV Generalization Challenge\footnote{Datasets are released at the SIGMORPHON Shared Task 2024 on Subword Tokenization. \url{https://github.com/sigmorphon/2024TokenST}}, aimed at evaluating the generalization abilities of models when faced with OOV words, and consisting of three subtasks: Word and Definition (WaD),  Word and Morphology (WaM), and Word and Word (WaW). More details in relation to the state-of-the-art generalization in NLP are available in Appendix~\ref{app:generalization}.

\subsection{Subtask 1: Word and Definition (WaD)}
\noindent \textbf{Description.} Classify whether a given word and a given definition match semantically. WaD examples can be seen in Table \ref{tab:WaD}.

\noindent \textbf{Dataset development.} We converted English Wordnet \cite{miller1990introduction}, a freely available lexico-semantic resource, into 207K word--definition pairs, annotating them with a `true' label. Then, we shuffled words and definitions between the instances and labeled them as `false'. To generate out-of-distribution (OOD) examples for development and test splits, we sampled words that were unseen during training, and for negative examples, we intentionally chose words that were lexically similar to the original words of given definitions, as illustrated by examples in Table~\ref{tab:WaD}. We sampled 10,500~instances for training, 1,500~instances for development, and 3,000~instances for testing.





\subsection{Subtask 2: Word and Morphology (WaM)}
\noindent \textbf{Description.} Classify whether a given word contains inflection, derivation, or compounding. The examples are shown in Table \ref{tab:WaM_example}.

\noindent \textbf{Dataset development.} The data for this subtask has been automatically converted from the SIGMORPHON Shared Task 2022 on Morpheme Segmentation \cite{batsuren-etal-2022-sigmorphon}. This shared task provided three morphological labels (inflection, derivation, compound) for 596K English Words. We first retrieved the most frequent 5000~subwords from the GPT-2 Tokenizer, which are shared across all splits (train, validation, test) and can be part of any subword composition in any split. The main evaluation setup is a compositional generalization \cite{hupkes2020compositionality} that every word in validation and test splits contained at least one unseen subword during training. These unseen subwords are outside of the most frequent 5000 subwords. The train, validation, and test splits contain 5400, 900, and 1800 instances, respectively. All splits contain balanced distributions of instances across the three morphological categories and output labels.

\subsection{Subtask 3: Word and Word (WaW)}
\noindent \textbf{Description.} Classify whether two given words are semantically related. Examples are shown in Table \ref{tab:WaW_example}.

\noindent \textbf{Dataset development.}  
For this subtask, we collected word pairs that are semantically related to each other from the English Wordnet. We retrieved 17.7M word pairs, including siblings (13M), hypernyms (4.3M), synonyms (304K), antonyms (20K), meronyms (71K), and substances (5.3K). Overall, we generated 5,389 instances for training, 582 instances for development, and 1,133 instances for test split. The positive instances were randomly sampled 
from the main pool of 17.7 million word pairs, and for the negative instances, we first created a list of unique words, then we generated negative samples by sampling two distinct words from the list for each instance and verified that each generated negative instance was absent from the word pair pool. To evaluate generalization abilities, words in the development and test splits are unseen on the training split. 

\begin{table*}[ht]
\small
\centering
\caption{\label{tab:waw_cov}WaW label coverage of test split by umLabeller categories.}
\newcolumntype{R}{>{\raggedleft\arraybackslash}X}
\newcolumntype{C}{>{\centering\arraybackslash}X}
\newcolumntype{L}{>{\raggedright\arraybackslash}X}
\begin{tabularx}{\textwidth}{l|CCCCCC|c}
Model   & vocab\&vocab & morph\&morph & alien\&alien & morph\&alien & vocab\&alien & vocab\&morph & total    \\
\hline
\hline
ALBERT  & 14.0             & 8.0              & 20.9           & 14.2           & 23.5           & 19.2           & 99.9  \\
BERT    & 13.1           & 4.3            & 28.7           & 9.0              & 26.2           & 18.6           & 99.9  \\
RoBERTa & 12.1           & 3.8            & 32.1           & 8.5            & 25.5           & 17.9           & 99.9  \\
DeBERTa & 35.3           & 7.1            & 10.4           & 16.3           & 18.6           & 12.1           & 99.9 \\
\hline
\end{tabularx}
\end{table*}

\begin{table*}[ht]
\small
\centering
\caption{\label{tab:waw}WaW test accuracies by umLabeller categories. All language models are their corresponding base models. All experiments trained on the best hyperparam sweep according to the validation results, averaged over five seeds.}
\newcolumntype{R}{>{\raggedleft\arraybackslash}X}
\newcolumntype{C}{>{\centering\arraybackslash}X}
\newcolumntype{L}{>{\raggedright\arraybackslash}X}
\begin{tabularx}{\textwidth}{l|cccccc|C}
Model   & vocab\&vocab & morph\&morph & alien\&alien & morph\&alien & vocab\&alien & vocab\&morph & total    \\
\hline
\hline
ALBERT  & 68.8$\pm$1.6       & 85.7$\pm$2.2       & 73.1$\pm$3.5       & 78.5$\pm$1.4       & 73.2$\pm$1.0       & 77.6$\pm$1.2       & 75.2$\pm$0.9 \\
BERT    & 73.2$\pm$2.2       & 89.8$\pm$1.6       & 77.0$\pm$1.5       & 79.9$\pm$1.7       & 75.6$\pm$1.2       & 77.5$\pm$1.6       & 77.3$\pm$0.7 \\
RoBERTa & 79.0$\pm$1.6       & 91.6$\pm$1.1       & 75.4$\pm$1.7       & 80.3$\pm$0.8       & 78.8$\pm$1.2       & 82.3$\pm$2.1       & 78.8$\pm$0.8 \\
DeBERTa & 80.7$\pm$0.5       & 91.4$\pm$1.0       & 85.3$\pm$1.4       & 83.9$\pm$0.9       & 77.1$\pm$1.1       & 85.0$\pm$1.0       & 82.3$\pm$0.5 \\
\hline
\hline
Average & 75.4$\pm$4.4       & 89.6$\pm$2.0       & 77.7$\pm$3.8       & 80.7$\pm$1.6       & 76.1$\pm$1.8       & 80.6$\pm$3.0       & - \\
\end{tabularx}
\end{table*}

\section{Manual validation}
\label{sec:validation}
We randomly sampled 300 words of UniMorph and its GPT-3 Tokenizer subword tokenization, half of them as morphological composition on umLabeller and the other half as alien. First, we instructed two computational linguists (recruited from this paper's authors) about definitions of morphological composition and alien composition; then, they were asked to validate the correctness of umLabeller given the subword compositions of the GPT-3 Tokenizer. The inter-agreement score of two validators was 99.3\% (Cohen's kappa is 0.829) and then in the conflicting examples, they were asked to discuss and reach a common agreement. The final accuracy of the sample was 98.0\%, and in 6 out of 300 samples, the UniMorph segmentations were incomplete, e.g., \textit{reform} instead of \textit{re- form}.  


\section{Experiments and Results}
\label{sec:results}
In this section, we present the empirical results of our OOV Generalization Challenge 1.0 benchmark on four publicly available models. 

\subsection{Experimental setup} We train the base and large variants of four models (BERT \cite{devlin2018bert}, RoBERTa \cite{liu2019roberta}, ALBERT-v2 \cite{lan2019albert}, DeBERTa-v3 \cite{he2021debertav3}), learning rates (1e-5, 2e-5, 3e-5), batch sizes (8, 16, 32) and epochs (3, 10) and select the best-performing model based on the validation set of the individual split.
Finally, we train five seeds of the best-performing model.
The main reason we selected these models is that they have different tokenizers, that vary in terms of the vocabulary size (30k, 50k and 128k) and the tokenization scheme (BPE, ULM).

\subsection{Task performances}
\begin{figure}[t]
\centering
\includegraphics[width=1\linewidth]{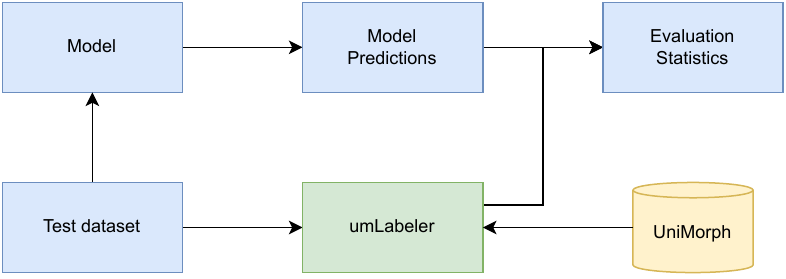}
\caption{UniMorph Labeller extends model evaluation to provide insights about the role of subword tokenization in terms of vocabulary words, alien composition, and morphological composition. Blue diagram is the common NLP pipeline.}
\label{fig:eval_pipeline}
\end{figure}

To understand how alien and morphological subword compositions impact the compositional generalizations of language models, we used umLabeller to divide the test splits of three subtasks into smaller groups, each of which represents one of umLabeller categories. This evaluation pipeline is illustrated in Figure \ref{fig:eval_pipeline}. 

\paragraph{Subtask 1: WaD results.} To inspect the language models in the WaD subtask, we divide its test split into three subgroups, each of which corresponds to the umLabeller categories of vocab, morph, and alien. Given the test dataset $\mathbb{D} = \{(x_i, y_i)\}_{i=1}^N$ and the tokenizer $\mathbb{T}$, each instance $x_i$ has two elements: word $w_i$ and definition $d_i$. Then the subgroup label is computed as $g_i = \textrm{umLabeller}(w_i, \mathbb{T}.\textrm{tokenize}(w_i))$. 

In Table \ref{tab:wad_results}, the WaD performances of language models are provided across all three umLabeller categories, along with the total accuracy. In subtask 1, if a given word $w$ belongs to the vocabulary of its tokenizer, then we have very high average accuracy (as minimum 93.2\%) across all language models. Overall, the vocab group in the test split has 95.0\% and 20.9\% absolute performance difference over morph and alien compositions, respectively. Thus, these results show that language models' generalization abilities are significantly better with morphological composition (by 5.4\% absolute accuracy) than alien composition. Since the vocab performances of all language models are significantly higher than those of other groups, the vocab coverage contributes heavily to the total accuracy. Therefore, it is not surprising to see the best-performing model, DeBERTa, in total accuracy.

\begin{table*}[ht]
\small
\centering
\caption{\label{tab:wic}WiC (word-in-context) development results by umLabeller categories. All language models are their corresponding base models. Tokenizer coverage represents the test split's coverage on the corresponding umLabeller category. All experiments trained on the best hyperparameter sweep according to the validation results, averaged over five seeds.}
\newcolumntype{R}{>{\raggedleft\arraybackslash}X}
\newcolumntype{C}{>{\centering\arraybackslash}X}
\newcolumntype{L}{>{\raggedright\arraybackslash}X}
\begin{tabularx}{\textwidth}{l|cc|ccc|CCC|C}
     & \multicolumn{2}{c|}{Tokenizer} & \multicolumn{3}{c|}{Dev distribution} & \multicolumn{4}{c}{Accuracy}             \\
\hline
\hline
Model   & type & size & vocab & morph & alien & \multicolumn{1}{c}{vocab}       & \multicolumn{1}{c}{morph}        & \multicolumn{1}{c|}{alien}       & total       \\
\hline
\hline
ALBERT  & ULM & 30k         & 80.3  & 6.1   & 10.8  & 72.3 $\pm$ 0.8 & 68.2 $\pm$ 1.6  & 62.3 $\pm$ 2.9 & 70.7 $\pm$ 0.7 \\
BERT    & BPE & 30k         & 77.9  & 4.1   & 15.2  & 71.6 $\pm$ 0.6 & 74.6 $\pm$ 2.8  & 59.3 $\pm$ 0.9 & 69.5 $\pm$ 0.4 \\
RoBERTa & BPE & 50k         & 80.7  & 3.8   & 12.7  & 67.6 $\pm$ 1.8 & 63.3 $\pm$ 7.3 & 62.4 $\pm$ 2.7 & 67.1 $\pm$ 2.1 \\
\end{tabularx}
\end{table*}

\paragraph{Subtask 2: WaM results.} As WaD subtask, we used the exact same settings to divide the WaM test split into three umLabeller subgroups as each instance $x_i$ has two elements: word $w_i$ and morphology $m_i$. Each morphology $m_i$ is one of three morphological types: inflection, derivation, and compound. 
The language model performances of the WaM subtask are described in Table \ref{tab:WaM}, showing all three umLabeller categories and all three morphologies. All four models performed better on the group with morphological composition over alien composition by large margins with absolute improvements 2.7\% (DeBERTa) $\rightarrow$ 7.2\% (RoBERTa).

\paragraph{Subtask 3: WaW results.}
Given the WaW test dataset $\mathbb{D} = \{(x_i, y_i)\}_{i=1}^N$ and the tokenizer $\mathbb{T}$, each instance $x_i$ has two words: word $a_i$ and word $b_i$ then two subgroup labels are produced as $g^a_i = \textrm{umLabeller}(w_i, \mathbb{T}.\textrm{tokenize}(a_i))$ and $g^b_i = \textrm{umLabeller}(b_i, \mathbb{T}.\textrm{tokenize}(b_i))$. Depending on values of $g^a_i$ and $g^b_i$, we divide the test dataset into six groups of umLabeller label pairs. The label pair list can be seen from the header row in Table \ref{tab:waw}. The subgroup coverage distributions of four tokenizers are shown in Table \ref{tab:waw_cov}. Table \ref{tab:waw} reports the test accuracies of four models across the six subgroups of the test split. Surprisingly, each model performed best in a group with two morphologically composed words, unlike groups with vocabulary words in the previous two subtasks. This result shows that OOV words are semantically more informative than vocabulary words, only in this subtask.


\paragraph{Comparison to WiC.}
We compare the results of our three generalization tasks to the state-of-the-art Word-in-Context (WiC) task mentioned in Section~\ref{sec:sota}. WiC takes one word and two contextual sentences in input, and the goal of the task is to predict whether the word has the same meaning in both contexts. Table \ref{tab:wic} shows the performances of three language models across three umLabeller categories of subword compositions of the given words in its development split. The WiC corpus demonstrates a considerable performance decrease for all three models on alien composition (10\% for ALBERT, 5.3\% for BERT, and 5.2\% for RoBERTa). 

\section{Conclusion}
\label{sec:conclusion}
In this paper, we first formulated notions of morphological and alien subword compositions. Thus, we argued that alien subword composition fundamentally differs from morphological composition in supporting the semantic compositionality of word meanings. To this end, we first presented an open-source, large-scale, high-quality inspection tool, umLabeller, to classify a given subword composition as alien or morphological. Then, we introduced the OOV Generalization Challenge 1.0 benchmark composed of three downstream subtasks. The empirical results in the experiments show that morphological composition supports better the semantic compositionality of OOV word meanings than alien composition. Overall, our findings suggest that the generalization capabilities of language models can be further improved if language models use morphological subword tokenization. Also, the text characterization tools for language models become foundational in NLP evaluation \cite{simig2022text,chen2024catastrophic} and  umLabeller can play a crucial role in inspecting the impact of subword tokenization in language models. 

\section*{Limitations}
Our work is limited to addressing tokenization in English. The correctness and coverage of umLabeller directly depend on UniMorph English morpheme segmentation data, which is freely available on GitHub. We are publicly welcoming everyone, interested in fixing the mistakes and increasing the coverage through GitHub pull requests.

\section*{Acknowledgments}
O.U. and Y.P. were supported by the Israel Science Foundation (grant No. 1166/23).

\bibliography{anthology,custom}

\appendix

\section{Appendix}

\subsection{umLabeller Algorithm}
\label{app:algorithm}
Here, we provide a minimal Python implementation of \textrm{umLabeller} in Algorithm \ref{umlabel-algorithm}.

\begin{algorithm}[H]
\begin{lstlisting}
unimorph = {'motivated': ['motive','ate','ed']}
merges = {'motive ate':'motivate', 'ate ed':'ated', 
'motivate ed':'motivated', 'motive ated':'motivated', 
'motive ate ed':'motivated', 'motive':'motive', 
'ate':'ate', 'ed':'ed', 'ated':'ated'}

def label(s, m):
  if len(s) > len(m): return 'alien'
  if len(s) == 2:
  	for i in range(len(m)-1):
    	if s[0] == merges(' '.join(m[:i+1])):
    		return 'morph'
    	if s[1] == merges(' '.join(m[i+1:])):
    		return 'morph'
    return 'alien'
  elif s[0] == m[0]:
    return label(s[1:], m[1:])
  elif s[-1] == m[-1]:
    return label(s[:-1], m[:-1])
  elif len(s) == len(m):
  	return 'alien'
  else:
  	a = label(s,m[:-2]+[merges[' '.join(m[-2:])]])
  	if a == 'morph':
  		return 'morph'
    return label(s,[merges[' '.join(m[:2])]]+m[2:])

def umLabeller(word, subwords):
  if len(subwords) == 1: return 'vocab'
  if word not in unimorph: return 'n.a'
  morphs = unimorph[word]
  return label(subwords, morphs)
\end{lstlisting}
\caption{UniMorph Labeller}
\label{umlabel-algorithm}
\end{algorithm}

\subsection{Relation to Generalization in NLP} 
\label{app:generalization}
According to the state-of-the-art generalization taxonomy in NLP \cite{Hupkes2023}, our benchmark has an cognitive and practical motivation, evaluating compositional and morphological generalization under fully-generated covariate shift between the finetuning and test stage as shown in Figure \ref{fig:genbench}.

\newcommand{\tabularwidth}{\columnwidth}

\newcommand{\expone}{$\square$}
\newcommand{\exptwo}{$\bigtriangleup$}
\newcommand{\expthree}{$\bigcirc$}

\begin{figure}[h]
\caption{Generalization Benchmark card}
\label{fig:genbench}
\centering
\renewcommand{\arraystretch}{1.1}         
\setlength{\tabcolsep}{0mm}         
\begin{tabular}{|p{\tabularwidth}<{\centering}|}         
\hline
               
\rowcolor{gray!60}               
\textbf{Motivation} \\               
\footnotesize
\begin{tabular}{p{0.25\tabularwidth}<{\centering} p{0.25\tabularwidth}<{\centering} p{0.25\tabularwidth}<{\centering} p{0.25\tabularwidth}<{\centering}}                        
\textit{Practical} & \textit{Cognitive} & \textit{Intrinsic} & \textit{Fairness}\\
\expone\hspace{0.8mm}		
& \expone\hspace{0.8mm}\exptwo\hspace{0.8mm}\expthree\hspace{0.8mm}		
& 		
& 		

\vspace{2mm} \\
\end{tabular}\\
               
\rowcolor{gray!60}               
\textbf{Generalization type} \\               
\footnotesize
\begin{tabular}{m{0.17\tabularwidth}<{\centering} m{0.20\tabularwidth}<{\centering} m{0.14\tabularwidth}<{\centering} m{0.17\tabularwidth}<{\centering} m{0.18\tabularwidth}<{\centering} m{0.14\tabularwidth}<{\centering}}                   
\textit{Compo- sitional} & \textit{Structural} & \textit{Cross Task} & \textit{Cross Language} & \textit{Cross Domain} & \textit{Robust- ness}\\
\expone\hspace{0.8mm}\expthree\hspace{0.8mm}		
& \exptwo\hspace{0.8mm}		
& 		
& 		
& 		
& 		

\vspace{2mm} \\
\end{tabular}\\
             
\rowcolor{gray!60}             
\textbf{Shift type} \\             
\footnotesize
\begin{tabular}{p{0.25\tabularwidth}<{\centering} p{0.25\tabularwidth}<{\centering} p{0.25\tabularwidth}<{\centering} p{0.25\tabularwidth}<{\centering}}                        
\textit{Covariate} & \textit{Label} & \textit{Full} & \textit{Assumed}\\  
\expone\hspace{0.8mm}\exptwo\hspace{0.8mm}\expthree\hspace{0.8mm}		
& 		
& 		
& 		

\vspace{2mm} \\
\end{tabular}\\
             
\rowcolor{gray!60}             
\textbf{Shift source} \\             
\footnotesize
\begin{tabular}{p{0.25\tabularwidth}<{\centering} p{0.25\tabularwidth}<{\centering} p{0.25\tabularwidth}<{\centering} p{0.25\tabularwidth}<{\centering}}                          
\textit{Naturally occuring} & \textit{Partitioned natural} & \textit{Generated shift} & \textit{Fully generated}\\
& 		
& 		
& \expone\hspace{0.8mm}\exptwo\hspace{0.8mm}\expthree\hspace{0.8mm}		

\vspace{2mm} \\
\end{tabular}\\
             
\rowcolor{gray!60}             
\textbf{Shift locus}\\             
\footnotesize
\begin{tabular}{p{0.25\tabularwidth}<{\centering} p{0.25\tabularwidth}<{\centering} p{0.25\tabularwidth}<{\centering} p{0.25\tabularwidth}<{\centering}}                         
\textit{Train--test} & \textit{Finetune train--test} & \textit{Pretrain--train} & \textit{Pretrain--test}\\
& \expone\hspace{0.8mm}\exptwo\hspace{0.8mm}\expthree\hspace{0.8mm}		
& 		
& 		

\vspace{2mm} \\
\end{tabular}\\

\hline
\end{tabular}
\end{figure}

\end{document}